# Modeling Chaotic Behavior of Stock Indices Using Intelligent Paradigms

**Ajith Abraham, Ninan Sajith Philip[1] and P. Saratchandran[2]**

Department of Computer Science, Oklahoma State University,
Tulsa, Oklahoma 74106, USA, Email: ajith.abraham@ieee.org

[1]Department of Physics, Cochin University of Science and Technology, India,
Email: nsp@stthom.ernet.in

[2]School of Electrical and Electronic Engineering, Nanyang Technological University,
Singapore 639798, E-mail: epsarat@ntu.edu.sg

## Abstract

The use of intelligent systems for stock market predictions has been widely established. In this paper, we investigate how the seemingly chaotic behavior of stock markets could be well represented using several connectionist paradigms and soft computing techniques. To demonstrate the different techniques, we considered Nasdaq-100 index of Nasdaq Stock Market$^{SM}$ and the S&P CNX NIFTY stock index. We analyzed 7 year's Nasdaq 100 main index values and 4 year's NIFTY index values. This paper investigates the development of a reliable and efficient technique to model the seemingly chaotic behavior of stock markets. We considered an artificial neural network trained using Levenberg-Marquardt algorithm, Support Vector Machine (SVM), Takagi-Sugeno neuro-fuzzy model and a Difference Boosting Neural Network (DBNN). This paper briefly explains how the different connectionist paradigms could be formulated using different learning methods and then investigates whether they can provide the required level of performance, which are sufficiently good and robust so as to provide a reliable forecast model for stock market indices. Experiment results reveal that all the connectionist paradigms considered could represent the stock indices behavior very accurately.

## 1. INTRODUCTION

Prediction of stocks is generally believed to be a very difficult task. The process behaves more like a random walk process and time varying. The obvious complexity of the problem paves way for the importance of intelligent prediction paradigms. During the last decade, stocks and futures traders have come to rely upon various types of intelligent systems to make trading decisions [1][3][7][11][18][19][26][23][28]. Several intelligent systems have in recent years been developed for modelling expertise, decision support and complicated automation tasks etc [28][9][15][5][24][16][29][4][17]. In this paper, we analysed the seemingly chaotic behaviour of two well-known stock indices namely Nasdaq-100 index of Nasdaq$^{SM}$ [21] and the S&P CNX NIFTY stock index [22].

Nasdaq-100 index reflects Nasdaq's largest companies across major industry groups, including computer hardware and software, telecommunications, retail/wholesale trade and biotechnology [21]. The Nasdaq-100 index is a modified capitalization-weighted index, which is designed to limit domination of the Index by a few large stocks while generally retaining the capitalization ranking of companies. Through an investment in Nasdaq-100 index tracking stock, investors can participate in the collective performance of many of the Nasdaq stocks that are often in the news or have become household names. Similarly, S&P CNX NIFTY is a well-diversified 50 stock index accounting for 25 sectors of the economy [22]. It is used for a variety of purposes such as benchmarking fund portfolios, index based derivatives and index funds. The CNX Indices are computed using market capitalisation weighted method, wherein the level of the Index reflects the total market value of all the stocks in the index relative to a particular base period. The method also takes into account constituent changes in the index and importantly corporate actions such as stock splits, rights, etc without affecting the index value.

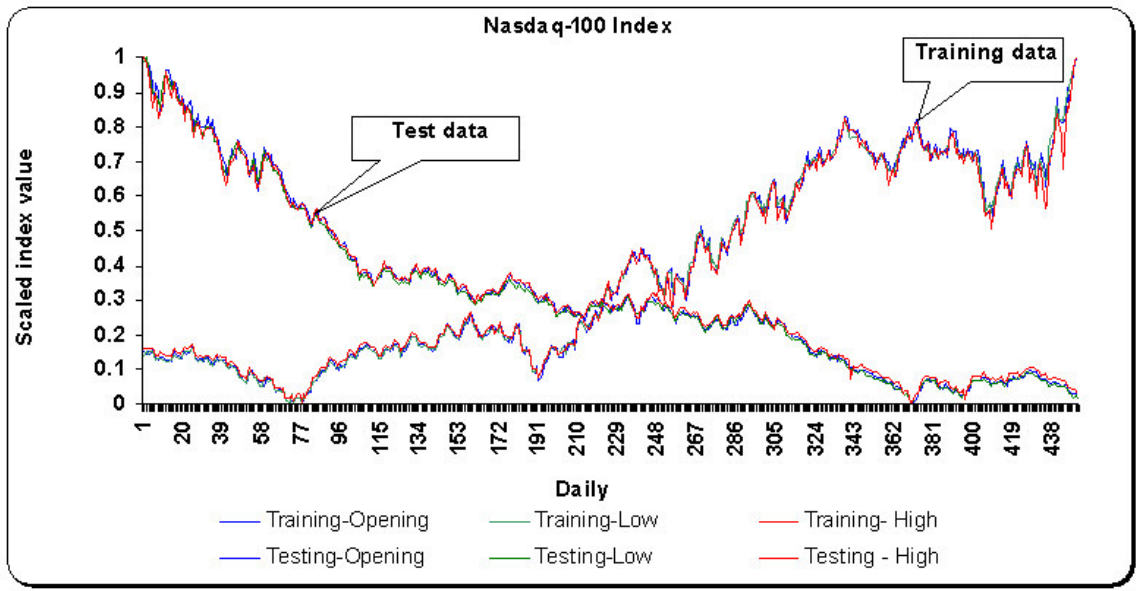

**Figure 1.** Training and Test data sets for Nasdaq-100 Index (b) NIFTY index

Our research is to investigate the performance analysis of four different connectionist paradigms for modelling the Nasdaq-100 and NIFTY stock market indices. The four different techniques considered are an artificial neural network trained using the Levenberg-Marquardt algorithm [6], support vector machine [27], difference boosting neural network [25] and a Takagi-Sugeno fuzzy inference system learned using a neural network algorithm (neuro-fuzzy model) [13]. Neural networks are excellent forecasting tools and can learn from scratch by adjusting the interconnections between layers. Support vector machines offer excellent learning capability based on statistical learning theory. Fuzzy inference systems are excellent for decision making under uncertainty. Neuro-fuzzy computing is a popular framework wherein neural network training algorithms are used to fine-tune the parameters of fuzzy inference systems. We analysed the Nasdaq-100 index value from 11 January 1995 to 11 January 2002 [21] and the

NIFTY index from 01 January 1998 to 03 December 2001 [22]. For both the indices, we divided the entire data into almost two equal parts. No special rules were used to select the training set other than ensuring a reasonable representation of the parameter space of the problem domain. The complexity of the training and test data sets for both indices are depicted in Figures 1 and 2 respectively. In Section 2 we briefly describe the different connectionist paradigms followed by experimentation setup and results in Section 3. Some conclusions are also provided towards the end.

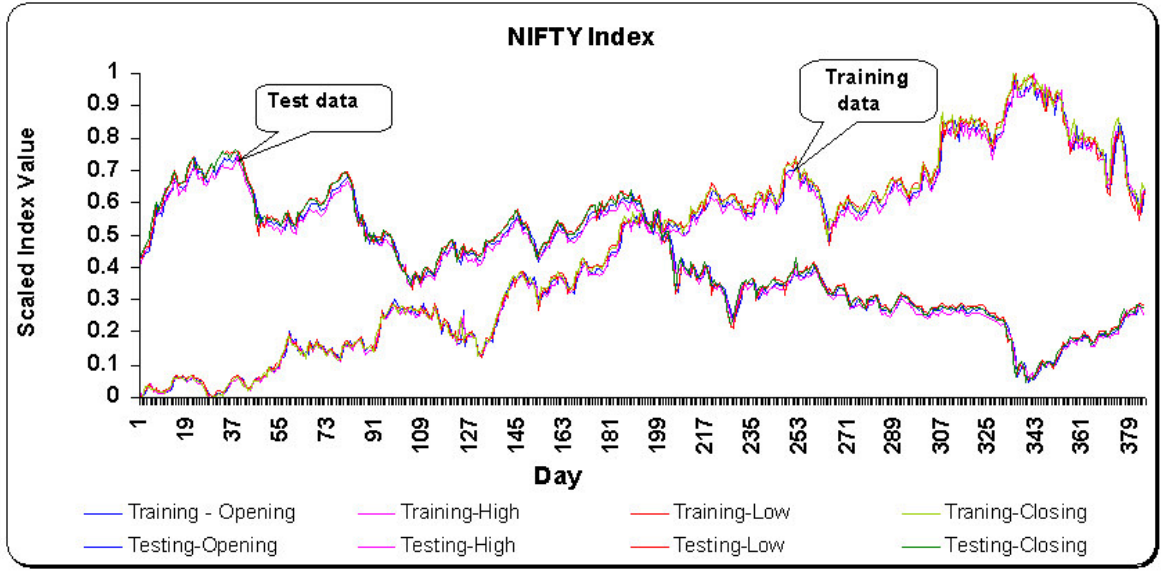

**Figure 2.** Training and Test data sets for NIFTY index

## 2. INTELLIGENT SYSTEMS: A CONNECTIONIST MODEL APPROACH

Connectionist models "learn" by adjusting the interconnections between layers. When the network is adequately trained, it is able to generalize relevant output for a set of input data. Learning typically occurs by example through training, where the training algorithm iteratively adjusts the connection weights (synapses). In an artificial neural network learning occurs by the iterative updating of connection weights using a learning algorithm.

### 2.1 ARTIFICIAL NEURAL NETWORKS

The artificial neural network (ANN) methodology enables us to design useful nonlinear systems accepting large numbers of inputs, with the design based solely on instances of input-output relationships. For a training set $T$ consisting of $n$ argument value pairs and given a $d$-dimensional argument $x$ and an associated target value $t$ will be approximated by the neural network output. The function approximation could be represented as

$T = \{(x_i, t_i) : i = 1 : n\}$

In most applications the training set $T$ is considered to be noisy and our goal is not to reproduce it exactly but rather to construct a network function that generalizes well to new function values. We will try to address the problem of selecting the weights to learn the training set. The notion of closeness on the training set $T$ is typically formalized through an error function of the form

$$\psi_T = \sum_{i=1}^{n} \|y_i - t_i\|^2 \tag{1}$$

where $y_i$ is the network output. Our target is to find a neural network $\eta$ such that the output $y_i = \eta(x_i, w)$ is close to the desired output $t_i$ for the input $x_i$ (w = strengths of synaptic connections). The error $\psi_T = \psi_T(w)$ is a function of $w$ because $y = \eta$ depends upon the parameters $w$ defining the selected network $\eta$. The objective function $\psi_T(w)$ for a neural network with many parameters defines a highly irregular surface with many local minima, large regions of little slope and symmetries. The common node functions (tanh, sigmoidal, logistic etc) are differentiable to arbitrary order through the chain rule of differentiation, which implies that the error is also differentiable to arbitrary order. Hence we are able to make a Taylor's series expansion in $w$ for $\psi_T$. We shall first discuss the algorithms for minimizing $\psi_T$ by assuming that we can truncate a Taylor's series expansion about a point $w^o$ that is possibly a local minimum. The gradient (first partial derivative) vector is represented by

$$g(w) = \nabla \psi_T \big|_w = \left[ \frac{\partial \psi_T}{\partial w_i} \right]_w \tag{2}$$

The gradient vector points in the direction of steepest increase of $\psi_T$ and its negative points in the direction of steepest decrease. The second partial derivative also known as Hessian matrix is represented by $H$

$$H(w) = H_{ij}(w) = \nabla^2 \psi_T(w) = \frac{\partial^2 \psi_T(w)}{\partial w_i \partial w_j} \tag{3}$$

The Taylor's series for $\psi_T$, assumed twice continuously differentiable about $w^0$, can now be given as

$$\psi_T(w) = \psi_T(w^0) + g(w^0)^T (w - w^0)^T + \frac{1}{2}(w - w^0)^T H(w^0)(w - w^0) \\ + O(\| w - w^0 \|^2) \tag{4}$$

where $O(\delta)$ denotes a term that is of zero-order in small $\delta$ such that $\lim_{\delta \to 0} \frac{O(\delta)}{\delta} = 0$. If for example there is continuous derivative at $w^0$, then the remainder term is of order $\| w - w^0 \|^3$ and we can reduce (4) to the following quadratic model

$$m(w) = \psi_T(w^0) + g(w^0)^T (w - w^0) + \frac{1}{2}(w - w^0)^T H(w^0)(w - w^0) \tag{5}$$

Taking the gradient in the quadratic model of (5) yields

$$\nabla m = g(w^0) + H(w - w^0) \tag{6}$$

If we set the gradient $g=0$ and solving for the minimizing $w^*$ yields

$$w^* = w^0 - H^{-1}g \quad (7)$$

The model $m$ can now be expressed in terms of minimum value of $w^*$ as

$$m(w^*) = m(w^0) + \frac{1}{2} g(w^0)^T H^{-1} g(w^0)$$

$$m(w) = m(w^*) + \frac{1}{2}(w - w^*)^T H(w^*)(w - w^*) \quad (8)$$

a result that follows from (5) by completing the square or recognizing that $g(w^*)=0$. Hence starting from any initial value of the weight vector, we can in the quadratic case move one step to the minimizing value when it exists. This is known as Newton's approach and can be used in the non-quadratic case where H is the Hessian and is positive definite.

### 2.1.1 LEVENBERG-MARQUARDT ALGORITHM

The Levenberg-Marquardt (LM) algorithm [6] exploits the fact that the error function is a sum of squares as given in (1). Introduce the following notation for the error vector and its Jacobian with respect to the network parameters $w$

$$J = J_{ij} = \frac{\partial e_j}{\partial w_i}, i = 1:p, j = 1:n \quad (9)$$

The Jacobian matrix is a large $p \times n$ matrix, all of whose elements are calculated directly by backpropagation technique. The $p$ dimensional gradient $g$ for the quadratic error function can be expressed as

$$g(w) = \sum_{i=1}^{n} e_i \nabla e_i(w) = Je$$

and the Hessian matrix by

$$H = H_{ij} = \frac{\partial^2 \psi_T}{\partial w_i \partial w_j} = \frac{1}{2}\sum_{k=1}^{n} \frac{\partial^2 e_k^2}{\partial w_i \partial w_j} = \sum_{k=1}^{n}\left(e_k \frac{\partial^2 e_k}{\partial w_i \partial w_j} + \frac{\partial e_k \partial e_k}{\partial w_i \partial w_j}\right)$$

$$= \sum_{k=1}^{n}\left(e_k \frac{\partial^2 e_k}{\partial w_i \partial w_j} + J_{ik}J_{jk}\right) \quad (10)$$

Hence defining $D = \sum_{i=1}^{n} e_i \nabla^2 e_i$ yields the expression

$$H(w) = JJ^T + D \quad (11)$$

The key to the LM algorithm is to approximate this expression for the Hessian by replacing the matrix $D$ involving second derivatives by the much simpler positively scaled unit matrix $\epsilon I$. The LM is a descent algorithm using this approximation in the form

$$M_k = \left[JJ^T + \in I\right]^{-1}, w_{k+1} = w_k - \alpha_k M_k g(w_k) \qquad (12)$$

Successful use of LM requires approximate line search to determine the rate $\alpha_k$. The matrix $JJ^T$ is automatically symmetric and non-negative definite. The typically large size of J may necessitate careful memory management in evaluating the product $JJ^T$. Hence any positive $\in$ will ensure that $M_k$ is positive definite, as required by the descent condition. The performance of the algorithm thus depends on the choice of $\in$.

When the scalar $\in$ is zero, this is just Newton's method, using the approximate Hessian matrix. When $\in$ is large, this becomes gradient descent with a small step size. As Newton's method is more accurate, $\in$ is decreased after each successful step (reduction in performance function) and is increased only when a tentative step would increase the performance function. By doing this, the performance function will always be reduced at each iteration of the algorithm.

## 2.2 SUPPORT VECTOR MACHINES (SVM)

Support Vector Machines (SVMs) [27] combine several techniques from statistics, machine learning and neural networks. SVM perform structural risk minimization. They create a classifier with minimized VC (Vapnik and Chervonenkis) dimension. If the VC Dimension is low, the expected probability of error is low as well, which means good generalization. SVM has the common capability to separate the classes in the linear way. However, SVM also has another specialty that it is using a linear separating hyperplane to create a classifier, yet some problems can't be linearly separated in the original input space. Then SVM uses one of the most important ingredients called kernels, i.e., the concept of transforming linear algorithms into nonlinear ones via a map into feature spaces. Figures 3 and 4 illustrate two categories of data using Y+ and Y- symbols.

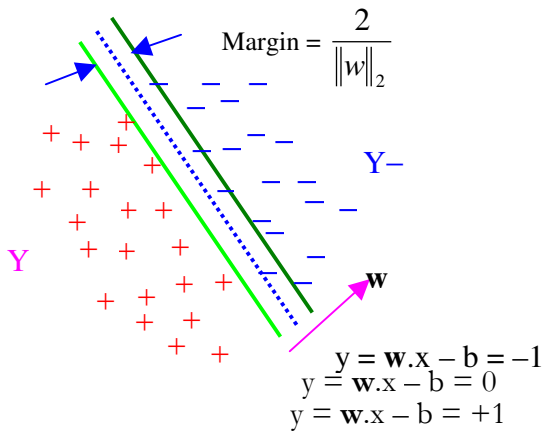
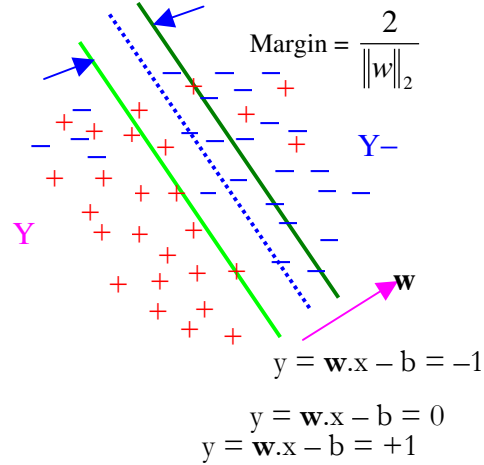

**Figure 3:** The linearly separable case.   **Figure 4:** The linearly inseparable case.

## 2.2.1 LINEAR SVM

We consider N training data points $\{(x_1, y_1), (x_2, y_2),.....,(x_N, y_N)\}$ where $x_i \in R^d$ and $y_i \in \{\pm 1\}$. We would like to explain a linear separating hyperplane classifier:

$$f(x) = \text{sgn}(w.x - b) \tag{13}$$

Furthermore, we want this hyperplane to have the maximum separating margin with respect to the two classes. Specifically, we want to find this hyperplane HP : y = w.x – b = 0 and two hyperplanes parallel to it and with equal distances to it,

$$HP_1 : y = w.x - b = +1 \text{ and } HP_2 : y = w.x - b = -1 \tag{14}$$

with the condition that there are no data points between $HP_1$ and $HP_2$, and the distance between $HP_1$ and $HP_2$ is maximized.

For any separating plane HP and the corresponding $HP_1$ and $HP_2$, we can always normalize the coefficients vector w so that $HP_1$ will be y = w.x – b = +1, and $HP_2$ will be y = w.x – b = –1.

Our aim is to maximize the distance between $HP_1$ and $HP_2$. So there will be some positive examples on $HP_1$ and some negative examples on $HP_2$. These examples are called support vectors because only they participate in the definition of the separating hyperplane, and other examples can be removed and/or moved around as long as they don't cross the planes $HP_1$ and $HP_2$.

Recall that the 2-D, the distance from a point $(x_0, y_0)$ to a line Ax+Bx+C = 0 is $\frac{|Ax_0 + By_0 + C|}{\sqrt{A^2 + B^2}}$. Similarly, the distance of a point on $HP_1$ to HP : w.x – b = 0 is $\frac{|w.x - b|}{\|w\|} = \frac{1}{\|w\|}$, and the distance between $HP_1$ and $HP_2$ is $\frac{2}{\|w\|}$. So, in order to maximize the distance, we should minimize $\|w\| = w^T w$ with the condition that there are no data points between $HP_1$ and $HP_2$ w.x – b ≥ +1, for positive example $y_i = +1$ and w.x – b ≥ -1, for negative example $y_i = -1$

These two condition can be combined into: $y_i(w.x - b) \geq 1$

Now the problem can be formulated as

$$\min_{w,b} \frac{1}{2} w^T w \text{ subject to } y_i(w.x - b) \geq 1 \tag{15}$$

This is a convex, quadratic programming problem (in w, b) in a convex set.

Introducing Lagrange multipliers $\alpha_1, \alpha_2, ....\alpha_n \geq 0$, we have the following Lagrangian:

$$L(w,b,\alpha) \equiv \frac{1}{2} w^T w - \sum_{i=1}^{N} \alpha_i y_i (w.x_i - b) + \sum_{i=1}^{N} \alpha_i. \tag{16}$$

## 2.4.2 NON LINEAR SVM

When the two classes are non-linearly distributed then SVM can transform the data points to another high dimensional space such that the data points will be linearly separable. Let the transformation be $\Phi(\cdot)$. In the high dimensional space, we solve

$$L_D \equiv \sum_{i=1}^{N} \alpha_i - \frac{1}{2} \sum_{i,j} \alpha_i \alpha_j y_i y_j \Phi(x_i) \cdot \Phi(x_j) \tag{17}$$

Suppose, in addition, $\Phi(x_i) \cdot \Phi(x_j) = k(x_i, x_j)$. That is, the dot product in that high dimensional space is equivalent to a kernel function of the input space. So, we need not be explicit about the transformation $\Phi(\cdot)$ as long as we know that the kernel function $k(x_i, x_j)$ is equivalent to the dot product of some other high dimensional space.

The Mercers's condition can be used to determine if a function can be used as a kernel function:

There exists a mapping $\Phi$ and an expansion

$$K(x, y) = \sum_{i} \Phi(x)_i \Phi(y)_i \tag{18}$$

if and only if, for any $g(x)$ such that $\int g(x)^2 dx$ is finite, then

$$\int K(x, y) g(x) g(y) dx dy \geq 0. \tag{19}$$

The foundations of SVM have been developed by Vapnik [27] and are gaining popularity due to many attractive features, and promising empirical performance. The possibility of using different kernels allows viewing learning methods like Radial Basis Function Neural Network (RBFNN) or multi-layer Artificial Neural Networks (ANN) as particular cases of SVM despite the fact that the optimized criteria are not the same [14]. While ANNs and RBFNN optimizes the mean squared error dependent on the distribution of all the data, SVM optimizes a geometrical criterion, which is the margin and is sensitive only to the extreme values and not to the distribution of the data into the feature space. The SVM approach transforms data into a feature space F that usually has a huge dimension. It is interesting to note that SVM generalization depends on the geometrical characteristics of the training data, not on the dimensions of the input space. Training a support vector machine (SVM) leads to a quadratic optimization problem with bound constraints and one linear equality constraint. Vapnik [27] shows how training a SVM for the pattern recognition problem leads to the following quadratic optimization problem

Minimize: $$W(\alpha) = -\sum_{i=1}^{l} \alpha_i + \frac{1}{2} \sum_{i=1}^{l} \sum_{j=1}^{l} y_i y_j \alpha_i \alpha_j k(x_i, x_j) \tag{20}$$

Subject to $$\sum_{i=1}^{l} y_i \alpha_i \tag{21}$$

$$\forall i : 0 \leq \alpha_i \leq C$$

Where $l$ is the number of training examples $\alpha$ is a vector of $l$ variables and each component $\alpha_i$ corresponds to a training example $(x_i, y_i)$. The solution of (1) is the vector

$\alpha^*$ for which (1) is minimized and (2) is fulfilled. We used the SVMTorch for simulating the SVM learning algorithm [10].

## 2.3 NEURO-FUZZY SYSTEM

Neuro Fuzzy (NF) computing is a popular framework for solving complex problems [2]. If we have knowledge expressed in linguistic rules, we can build a Fuzzy Inference System (FIS) [8], and if we have data, or can learn from a simulation (training) then we can use ANNs. For building a FIS, we have to specify the fuzzy sets, fuzzy operators and the knowledge base. Similarly for constructing an ANN for an application the user needs to specify the architecture and learning algorithm. An analysis reveals that the drawbacks pertaining to these approaches seem complementary and therefore it is natural to consider building an integrated system combining the concepts. While the learning capability is an advantage from the viewpoint of FIS, the formation of linguistic rule base will be advantage from the viewpoint of ANN.

Figure 5 depicts the 6- layered architecture of multiple output ANFIS and the functionality of each layer is as follows:

***Layer-1.*** Every node in this layer has a node function. $O_i^1 = \mu_{A_i}(x)$, for i =1, 2 or $O_i^1 = \mu_{B_{i-2}}(y)$, for i=3,4,.... $O_i^1$ is the membership grade of a fuzzy set A ( = $A_1$, $A_2$, $B_1$ or $B_2$) and it specifies the degree to which the given input $x$ (or $y$) satisfies the quantifier A. Usually the node function can be any parameterized function. A gaussian membership function is specified by two parameters $c$ (membership function center) and $\sigma$ (membership function width).

guassian (x, c, $\sigma$) = $e^{-\frac{1}{2}\left(\frac{x-c}{\sigma}\right)^2}$. Parameters in this layer are referred to premise parameters.

***Layer-2.*** Every node in this layer multiplies the incoming signals and sends the product out. Each node output represents the firing strength of a rule.

$O_i^2 = w_i = \mu_{A_i}(x) \times \mu_{B_i}(y), i = 1,2......$, In general any T-norm operator that perform fuzzy "AND" can be used as the node function in this layer.

***Layer-3.*** The rule consequent parameters are determined in this layer.

$O_i^3 = f_i = xp_i + yq_i + r_i$, where $\{p_i, q_i, r_i\}$ are the rule consequent parameters.

***Layer-4.*** Every node i in this layer is with a node function

$O_i^4 = \sum \overline{w_i} f_i = \sum \overline{w_i}(p_i x + q_i y + r_i)$, where $\overline{w_i}$ is the output of layer 2

***Layer-5.*** Every node in this layer aggregates all the firing strengths of rules

$$O_i^5 = \sum_i \overline{w_i} \qquad (22)$$

***Layer-6.*** Every *i*-th node in this layer calculates the individual outputs.

$$O_i^6 = \overline{Output} = \frac{\sum \overline{w_i} f_i}{\sum_i \overline{w_i}}, i = 1,2..... \qquad (23)$$

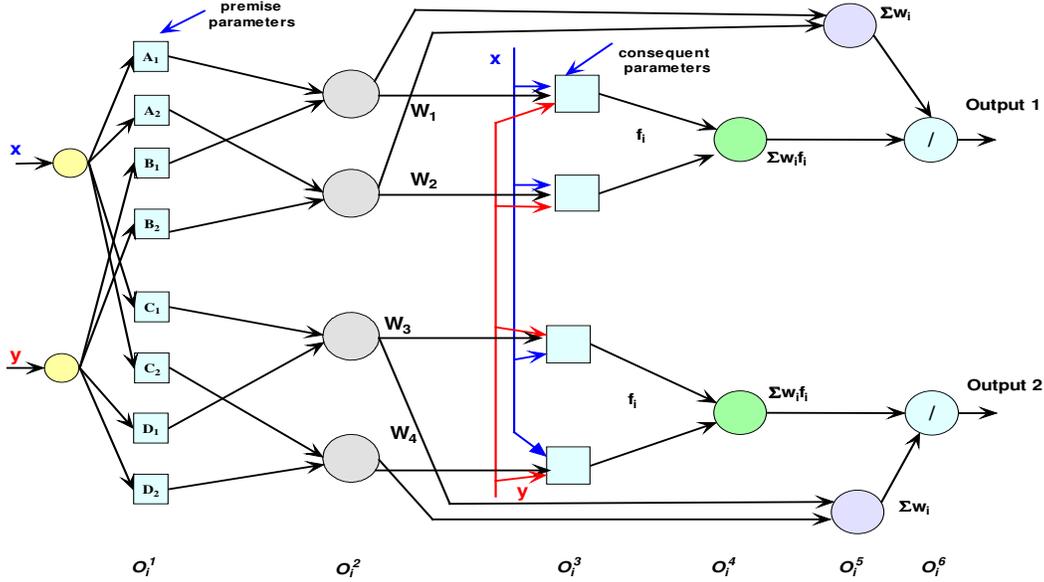

**Figure 5.** Architecture of ANFIS with multiple outputs

ANFIS uses a hybrid learning rule with a combination of gradient descent and least squares estimate [13]. Assuming a single output ANFIS represented by

$$output = F(\vec{I}, S) \qquad (24)$$

where $I$ is the set of input variables and $S$ is the set of parameters, if there exist a function H such that the composite function $H \circ F$ is linear in some of the elements of $S$, then these elements can be identified by the least squares method [13]. More formally, the parameter set $S$ can be decomposed into two sets:

$$S = S_1 \oplus S_2 \text{ (where } \oplus \text{ represents direct sum)}, \qquad (25)$$

such that $H \circ F$ is linear in the elements of $S_2$. Then upon applying $H$ to equation (6.1), we have:

$$H(output) = H \circ F(\vec{I}, S) \qquad (26)$$

which is linear in the elements of $S_2$. Now the given values of elements of $S_1$, we can plug $P$ training data sets into (6.3), and obtain a matrix equation:

$AX = B$ ($X$ = unknown vector whose elements are parameters in $S_2$) $\qquad (27)$

If $|S_2|=M$, ($M$= number of linear parameters) then the dimensions of $A$, $X$ and $B$ are $P \times M$, $M \times 1$ and $P \times 1$ respectively. Since $P$ is always greater than $M$, there is no exact solution to equation *(6.4)*. Instead a Least Square Estimate (LSE) of X, $X^*$, is sought to minimize the squared error $\|AX - B\|^2$. $X^*$ is computed using the pseudo-inverse of X:

$$X^* = (A^T A)^{-1} A^T B \qquad (28)$$

where $A^T$ is the transpose of A and $(A^T A)^{-1} A^T$ is the pseudo-inverse of A where $A^T A$ is non-singular. Due to computational complexity, in ANFIS a sequential method is deployed as follows:

Let the *i-th* row vector of matrix A defined in equation 6.4 be $a_i^T$ and *i-th* element of matrix B defined be $b_i^T$, then X can be calculated iteratively using the following sequential formulae:

$$\begin{aligned} X_{i+1} &= X_i + S_{i+1} a_{i+1} (b_{i+1}^T - a_{i+1}^T X_i) \\ S_{i+1} &= S_i - \frac{S_i a_{i+1} a_{i+1}^T S_i}{1 + a_{i+1}^T S_i a_{i+1}}, \quad i = 0,1,........, P-1 \end{aligned} \qquad (29)$$

where $S_i$ is often called the covariance matrix and the least squares estimate $X^*$ is equal to $X_P$. The initial condition to bootstrap (6.6) are $X_O=0$ and $S_O=\gamma I$, where $\gamma$ is a positive large number and $I$ is the identity matrix of dimension $M \times M$. For a multi output ANFIS, (6.6) is still applicable except the $output = F(\vec{I}, S)$ will become a column vector. Each epoch of this hybrid learning procedure is composed of a forward pass and a backward pass. In the forward pass, we have to supply the input data and functional signals go forward to calculate each node output until the matrices *A* and *B* in (6.4) are obtained, and the parameters in $S_2$ are identified by the sequential least squares formulae given in (6.6). After identifying parameters in $S_2$, the functional signals keep going forward till the error measure is calculated. In the backward pass, the error rates propagate from the output layer to the input layers, and the parameters in $S_1$ are updated by the gradient method given by

$$\Delta\alpha = -\eta \frac{\partial E}{\partial \alpha} \qquad (30)$$

where $\alpha$ is the generic parameter, $\eta$ is a learning rate and $E$ the error measure. For given fixed values of parameters in $S_1$, the parameters in $S_2$ thus found are guaranteed to be the global optimum point in the $S_2$ parameter space due to the choice of the squared error measure.

The procedure mentioned above is mainly for offline learning version. However, the procedure can be modified for an online version by formulating the squared error measure as a weighted version that gives higher weighting factors to more recent data pairs. This amounts to the addition of a forgetting factor λ to (29).

$$X_{i+1} = X_i + S_{i+1}a_{i+1}(b_{i+1}^T - a_{i+1}^T X_i)$$

$$S_{i+1} = \frac{1}{\lambda}\left[S_i - \frac{S_i a_{i+1} a_{i+1}^T S_i}{\lambda + a_{i+1}^T S_i a_{i+1}}\right] \quad i = 0,1,\ldots,P-1 \tag{31}$$

The value of $\lambda$ is between 0 and 1. The smaller the $\lambda$ is, faster the effects of old data decay. However, a smaller $\lambda$ sometimes causes numerical instability and should be avoided.

### 2.4 DIFFERENCE BOOSTING NEURAL NETWORK (DBNN)

DBNN is based on the Bayes principle that assumes the clustering of attribute values while boosting the attribute differences [25]. Boosting is an iterative process by which the network places emphasis on misclassified examples in the training set until it is correctly classified. The method considers the error produced by each example in the training set in turn and updates the connection weights associated to the probability $P(U_m/C_k)$ of each attribute of that example ($U_m$ is the attribute value and $C_k$ a particular class in $k$ number of different classes in the dataset). In this process, the probability density of identical attribute values flattens out and the differences get boosted up. Instead of the serial classifiers used in the AdaBoost algorithm, DBNN approach uses the same classifier throughout the training process. An error function is defined for each of the miss classified examples based on it distance from the computed probability of its nearest rival. The enhancement to the attribute is done such that the error produced by each example decides the correction to its associated weights. Since it is likely that more than one class would be sharing at least some of the same attribute values, this would lead to competitive update of their attribute weights, until either the classifier figures out the correct class or the number of iterations are completed. The net effect of this would be that the classifier would become more and more dependent on the differences in the examples rather than their similarities.

### 3. EXPERIMENTATION SETUP AND RESULTS

We considered 7 year's months stock data for Nasdaq-100 Index and 4 year's for NIFTY index. Our target is to develop efficient forecast models that could predict the index value of the following trade day based on the opening, closing and maximum values of the same on a given day. The training and test patterns for both the indices (scaled values) are illustrated in Figures 1 and 2. For the Nasdaq-100index the data sets were represented by the '*opening value*', '*low value*' and '*high value*'. NIFTY index data sets were represented by '*opening value*', '*low value*', '*high value*' and '*closing value*'. We used the *s*ame training and test data sets to evaluate the different connectionist models. More details are reported in the following sections. Experiments were carried out on a Pentium IV, 1.5 GHz Machine with 256 MB RAM and the codes were executed using MATLAB (ANN, ANFIS) and C++ (SVM, DBNN). Test data was presented to the trained connectionist network and the output from the network was compared with the actual index values in the time series.

The assessment of the prediction performance of the different connectionist paradigms were done by quantifying the prediction obtained on an independent data set. The maximum absolute percentage error (*MAP*) and mean absolute percentage error (*MAPE*) were used to study the performance of the trained forecasting model for the test data.

*MAP* is defined as follows:

$$MAP = max\left(\frac{|P_{actual,\ i} - P_{predicted,\ i}|}{P_{predicted,\ i}} \times 100\right),$$ where $P_{actual,\ i}$ is the actual index value on day *i* and $P_{predicted,\ i}$ is the forecast value of the index on that day. Similarly *MAPE* is given as

$$MAPE = \frac{1}{N}\sum_{i=1}^{N}\left[\frac{P_{actual,i} - P_{predicted,i}}{P_{actual,i}}\right] \times 100,$$ where *N* represents the total number of days.

- **ANN – LM algorithm**

We used a feedforward neural network with 4 input nodes and a single hidden layer consisting of 26 neurons. We used tanh-sigmoidal activation function for the hidden neurons. The training was terminated after 50 epochs and it took about 4 seconds to train each dataset.

- **Neuro-fuzzy training**

We used 3 triangular membership functions for each of the input variable and the 27 *if-then* fuzzy rules were learned for the Nasdaq-100 index and 81 *if-then* fuzzy rules for the NIFTY index. Training was terminated after 12 epochs and it took about 3 seconds to train each dataset.

- **Support Vector Machines and Difference Boosting Neural Network**

Both SVM and DBNN took less than I second to learn the two data sets.

- **Performance and Results Achieved**

Table 1 summarizes the training and test results achieved for the two stock indices using the four different approaches. Figures 3 and 4 depict the test results for the one day ahead prediction of Nasdaq-100 index and NIFTY index respectively.

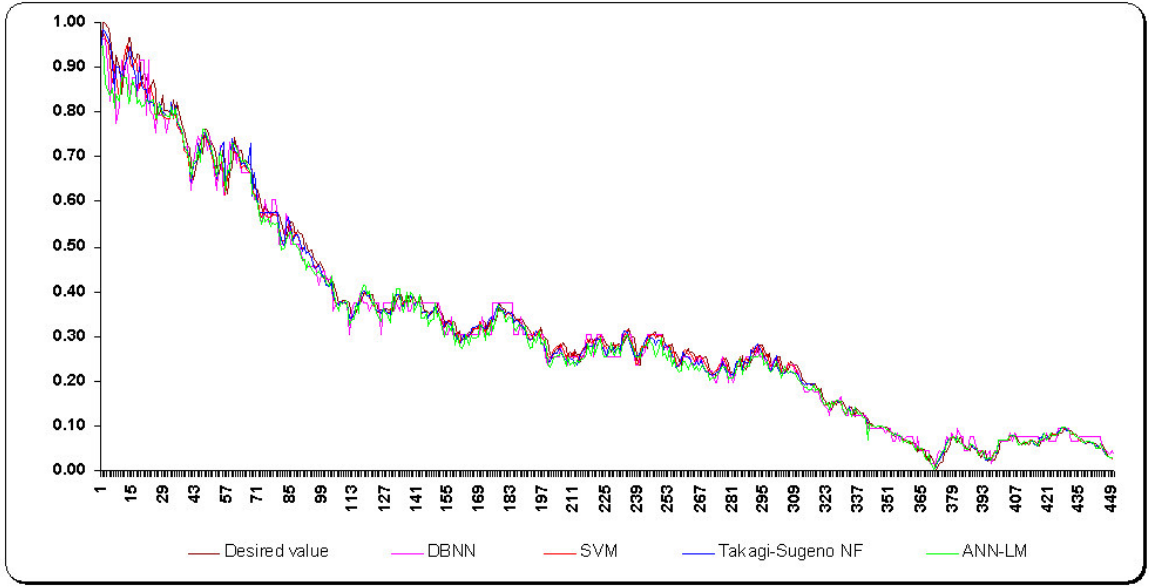

**Figure 3.** Test results showing the performance of the different methods for modeling Nasdaq-100 index

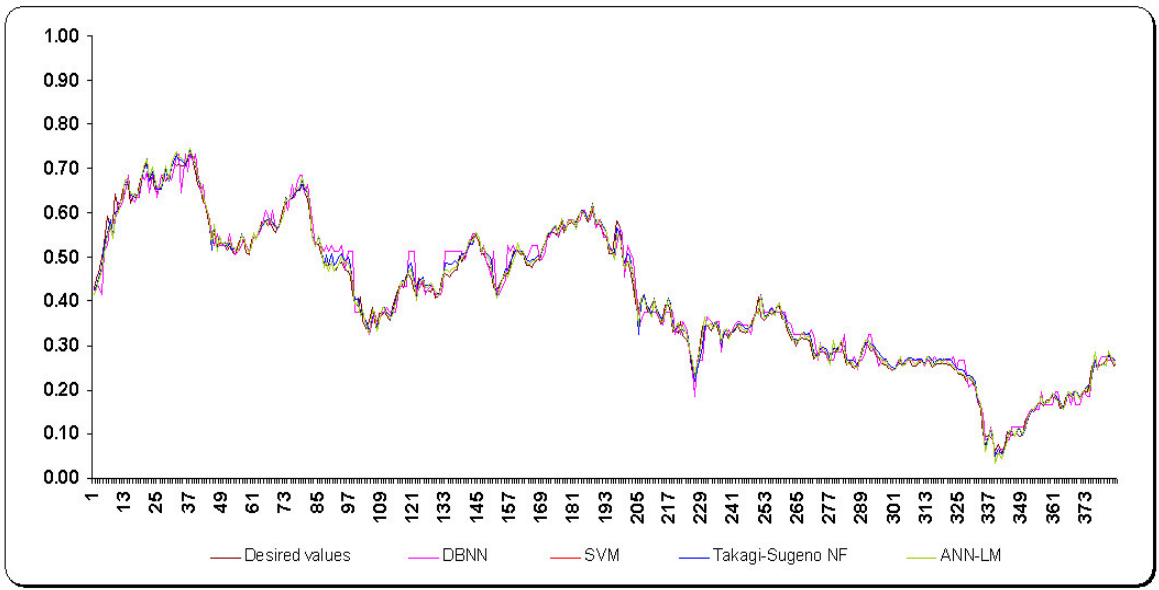

**Figure 4.** Test results showing the performance of the different methods for modeling NIFTY index

Table 1: Empirical comparison (training and test) of four learning methods

|  | SVM | Neuro-Fuzzy | ANN-LM | DBNN |
|---|---|---|---|---|
|  | Training results (RMSE) | | | |
| Nasdaq-100 | 0.02612 | 0.02210 | 0.02920 | 0.02929 |
| NIFTY | 0.01734 | 0.01520 | 0.01435 | 0.0174 |
|  | Testing results (RMSE) | | | |
| Nasdaq-100 | 0.01804 | 0.01830 | 0.02844 | 0.02864 |
| NIFTY | 0.01495 | 0.01271 | 0.01227 | 0.02252 |

Table 2: Statistical analysis of four learning methods (test data)

|  | SVM | Neuro-Fuzzy | ANN-LM | DBNN |
|---|---|---|---|---|
|  | Nasdaq-100 | | | |
| Correlation coefficient | 0.9977 | 0.9976 | 0.9955 | 0.9940 |
| MAP | 481.502 | 520.842 | 481.717 | 116.987 |
| MAPE | 7.170 | 7.615 | 9.032 | 9.429 |
|  | NIFTY | | | |
| Correlation coefficient | 0.9968 | 0.9967 | 0.9968 | 0.9890 |
| MAP | 72.53 | 40.37 | 73.94 | 37.99 |
| MAPE | 4.416 | 3.320 | 3.353 | 5.086 |

## 4. CONCLUSIONS

In this paper, we have demonstrated how the chaotic behavior of stock indices could be well represented by connectionist paradigms. Empirical results on the two data sets using four different models clearly reveal the efficiency of the proposed techniques. In terms of RMSE values, for Nasdaq-100 index, SVM performed marginally better than other models and for NIFTY index, ANN-LM approach gave the lowest generalization RMSE values. For both data sets, SVM has the lowest training time. For Nasdaq-100 index

SVM has the highest correlation coefficient and lowest value of MAPE but the lowest MAP value was for DBNN. Highest correlation coefficient was shared by SVM and ANN-LM approach for NIFTY index but the lowest MAP value was for the neuro-fuzzy approach. It is interesting to note that for predicting both index values, DBNN has the lowest MAP value.

Our research was to predict the share price for the following trade day based on the opening, closing and maximum values of the same on a given day. Our experimentation results indicate that the most prominent parameters that affect share prices are their immediate opening and closing values. The fluctuations in the share market are chaotic in the sense that they heavily depend on the values of their immediate forerunning fluctuations. Long-term trends exist, but are slow variations and this information is useful for long-term investment strategies. Our study focus on short term, on floor trades, in which the risk is higher. However, the results of our study show that even in the seemingly random fluctuations, there is an underlying deterministic feature that is directly enciphered in the opening, closing and maximum values of the index of any day making predictability possible.

Empriical results also shows that there are various advantages and disadvantages for the different techniques considered. Our future research will be oriented towards determining the optimal way to combine the different intelligent systems using an ensemble approach [12] so as to compliment the advantages and disadvantages of the different paradigms considered.